\documentclass[numsec,webpdf,modern,large]{oup-authoring-template}

\usepackage{color}
\usepackage{xcolor}
\usepackage{multirow}
\usepackage{graphicx}
\usepackage{booktabs}
\usepackage[switch]{lineno}

\graphicspath{{Fig/}}

\theoremstyle{thmstyleone}%
\theoremstyle{thmstyletwo}%
\theoremstyle{thmstylethree}%

\begin{document}

\journaltitle{Bioinformatics}
\DOI{DOI added during production}
\copyrightyear{YEAR}
\pubyear{YEAR}
\vol{XX}
\issue{x}
\access{Published: Date added during production}
\appnotes{Paper}

\firstpage{1}


\title[Short Article Title]{MolPIF: A Parameter Interpolation Flow Model for Molecule Generation}

\author[1,$\dagger$]{Yaowei Jin}
\author[1,2,$\dagger$]{Junjie Wang}
\author[5]{Yufan Tang}
\author[1]{Wenkai Xiang}
\author[3]{Duanhua Cao}
\author[3]{Dan Teng}
\author[3]{Zhehuan Fan}
\author[3]{Jiacheng Xiong}
\author[3]{Xia Sheng}
\author[3]{Chuanlong Zeng}
\author[1]{Duo An}

\author[3]{Mingyue Zheng}
\author[1,4]{Shuangjia Zheng}
\author[1,$\ast$]{Qian Shi\ORCID{0000-0002-7050-4314}}

\address[1]{\orgname{Lingang Laboratory}, \orgaddress{\state{Shanghai}, \postcode{200031}, \country{China}}}
\address[2]{\orgdiv{School of Information Science and Technology}, \orgname{ShanghaiTech University}, \orgaddress{\state{Shanghai}, \postcode{201210}, \country{China}}}
\address[3]{\orgdiv{Drug Discovery and Design Center, State Key Laboratory of Drug Research},  
    \orgname{Shanghai Institute of Materia Medica, Chinese Academy of Sciences},
    \orgaddress{
        \street{555 Zuchongzhi Road},
        \state{Shanghai},
        \postcode{201203},
        \country{China}
    }
}
\address[4]{\orgdiv{Global Institute of Future Technology}, \orgname{Shanghai Jiao Tong University}, \orgaddress{\state{Shanghai}, \postcode{200240}, \country{China}}}
\address[5]{\orgdiv{College of Computer Science and Artificial Intelligence}, \orgname{Fudan University}, \orgaddress{\state{Shanghai}, \postcode{200433}, \country{China}}}

\corresp[$\dagger$]{Yaowei Jin and Junjie Wang contributed equally.}

\corresp[$\ast$]{Corresponding author. Qian Shi. E-mail: \href{email:shiqian@lglab.ac.cn}{shiqian@lglab.ac.cn}}

\received{Date}{0}{Year}
\revised{Date}{0}{Year}
\accepted{Date}{0}{Year}


\abstract{
\textbf{Motivation:} Structure-based drug design (SBDD) has advanced with deep generative models, but bridging the gap between continuous atomic coordinates and discrete atom types remains a challenge. Current approaches, such as diffusion and flow matching models, often fail to unify these heterogeneous modalities, relying on separate strategies or ill-fitting Euclidean metrics for discrete variables. This lack of a consistent framework limits generative models' ability to capture the geometric and chemical structure of protein-ligand complexes.\\
\textbf{Results:} We present MolPIF, a parameter interpolation flow mechanism designed to unify the generation of continuous and discrete molecular variables. Unlike traditional flow models that operate in sample space, MolPIF interpolates between distributions in the parameter space, theoretically recovering Wasserstein-2 optimal transport for continuous coordinates and establishing Fisher–Rao geodesics for discrete atom types. We further incorporate a geometry-enhanced learning strategy to improve the capture of atomic contexts. Extensive evaluations on the CrossDocked2020 dataset demonstrate that MolPIF outperforms baselines in binding affinity, chemical validity, geometric fidelity and chemical space coverage. Additionally, MolPIF exhibits versatility in lead optimization and offers flexible prior distribution selection (such as Laplace), establishing a robust paradigm for SBDD.\\
\textbf{Availability:} Source code is freely available at \url{https://github.com/BLEACH366/MolPIF}.\\
\textbf{Contact:} \href{shiqian@lglab.ac.cn}{shiqian@lglab.ac.cn}\\
\textbf{Supplementary information:} Supplementary data are available at \textit{Bioinformatics}
online.}


\maketitle


\section{Introduction}
Computer-aided drug design (CADD) is pivotal to drug discovery, spanning from target validation to preclinical evaluation \citep{ragoza2022generating}. While structure-based drug design (SBDD) has demonstrated remarkable effectiveness in identifying lead compounds \citep{batool2019structure}, traditional CADD still struggles with imbalanced datasets and the exhaustive exploration of vast chemical and conformational spaces. To address this, AI-driven molecular generation—inspired by successes in AIGC \citep{guo2025deepseek, liu2025cachequant, tian2025audiox}—has emerged \citep{jiang2021interactiongraphnet, zang2020moflow}. These frameworks extract deep chemical insights from crystallographic data to navigate uncharted chemical territories, significantly accelerating the discovery of novel molecular structures.

Three-dimensional (3D) generative models have significantly advanced SBDD by incorporating protein pocket constraints and enabling end-to-end automation \citep{xie2022advances}. 
For SBDD tasks in 3D space, the mainstream approaches primarily leverage autoregressive or diffusion-based generative frameworks \citep{zhang2023molecule, gu2024aligning}.
However, these frameworks face distinct limitations: autoregressive models often suffer from mode collapse on unordered molecular data \citep{han2025infinity}, while diffusion and flow matching (FM) models struggle with the multimodal nature of atomic features \citep{ho2020denoising, song2019generative, lipman2022flow}. Specifically, molecular structures comprise heterogeneous variables—discrete (atom types), integer (formal charges), and continuous (spatial coordinates)—that require fundamentally different modeling methods \citep{song2023equivariant}. 
The prevailing strategy involves extending diffusion and FM models to discrete domains, employing advanced modeling techniques in order to achieve ideal performance \citep{cremer2025flowrflowmatchingstructureaware}.
Since the introduction of Bayesian Flow Networks (BFNs)\citep{graves2023bayesian} attempt to bridge this gap via parameter-space updates, they still model modalities separately. Consequently, the absence of a theoretically unified generative framework results in the persistent adverse effects of heterogeneous modalities on the quality of generated molecules.

To effectively unify discrete and continuous modeling, we introduce Parameter Interpolation Flow (PIF). By treating data as a superposition of Diracs, PIF interpolates distributional parameters between the data and a flexible prior. The model is optimized via Kullback-Leibler (KL) divergence to predict time-dependent parameters, enabling iterative refinement from the prior during inference. 
By shifting trajectories from sample to parameter space, PIF is highly versatile for both continuous and discrete data, allowing flexible prior selection and adaptation to various tasks without the need for complex closed-form derivations. 
PIF overcomes classical FM limitations: recovers 2-Wasserstein($W_2$)-optimal transport for Gaussian priors and extends this optimality to other distributions. For discrete data with an exponential family prior, PIF leverages Fisher–Rao geometry, ensuring probability paths evolve along exponential geodesics, accurately capturing the data manifold’s structure.

Molecular Parameter Interpolation Flow (MolPIF) extends the PIF framework to molecular generation by learning the parameter spaces associated with atomic coordinates (modeled as Gaussian distributions) and atomic types (modeled as Dirichlet distributions). During training, we incorporate a geometry-enhanced learning strategy, inspired by prior work \citep{cui2024geometry, peng2024atom}, to provide atomic-level contextual information of ligands to the model. This approach involves randomly masking subsets of atoms during training, enabling the model to dynamically optimize arbitrary atomic arrangements within a given molecular structure. As a result, MolPIF achieves superior performance in overall quality of the generated molecules.

Empirical evaluations conducted on the CrossDocked2020 dataset \citep{francoeur2020three} demonstrate that MolPIF has comprehensive advantages across five key dimensions: (1) Advanced de novo generation capability, producing molecules with enhanced binding properties and chemical validity; (2) Accurate geometric reproduction of molecular structural distributions, including rings, bond lengths, and bond angles; (3) Comprehensive chemical space modeling, with substantial coverage of 2D structural features and accurate 3D conformational distribution, while extending shape diversity beyond the reference distribution; (4) Flexible adaptation to different prior distributions, where we systematically compared and analyzed the advantages and limitations of using Gaussian versus Laplace distributions as priors for modeling atomic coordinates; (5) Effective lead optimization, demonstrating robust performance in enhancing drug candidate properties.

\section{Methods}\label{sec2}
\begin{figure*}[!t]%
\centering
\includegraphics[width=1\textwidth]{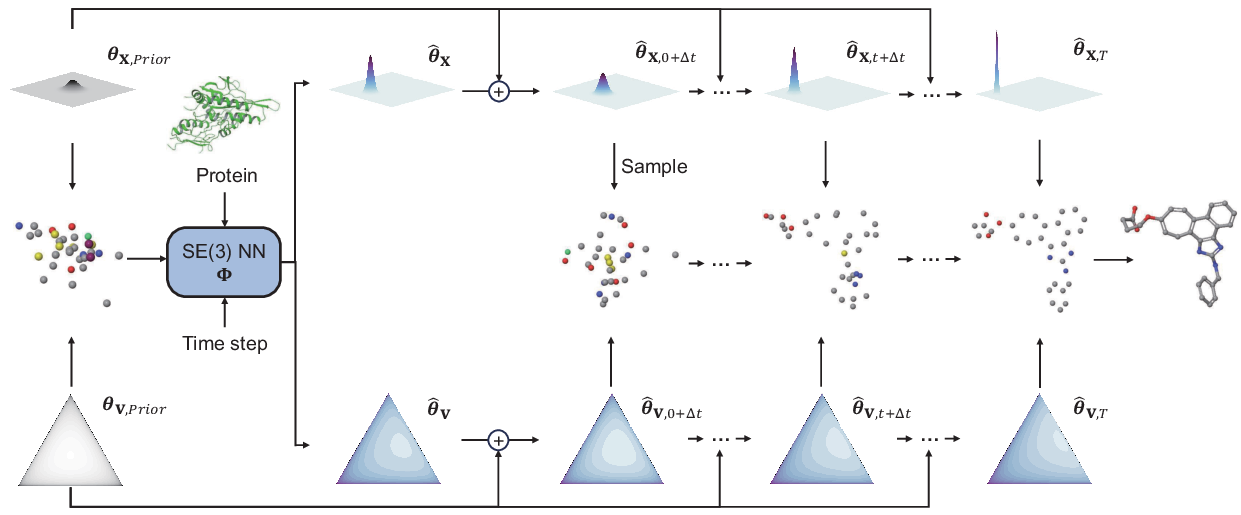}
\caption{
\textbf{The sampling process of the MolPIF framework.} 
Atomic coordinates $\mathbf{x}$ and types $\mathbf{v}$ are iteratively sampled and refined. An SE(3)-equivariant network updates distribution parameters by interpolating predictions with priors. At $t=1$, final molecular structures are sampled from the predicted Dirac distributions and assembled via OpenBabel.
}
\label{fig1}
\end{figure*}

\subsection{Definitions and notations}\label{subsec4}
Molecule generation based on receptor structure can be formulated as a conditional generation task. The input is a protein binding site \( \mathcal{P} = \{(\textbf{x}_P^{(i)}, \textbf{v}_P^{(i)})\}_{i=1}^{N_P} \), which contains \( N_P \) atoms with each \( \textbf{x}_P^{(i)} \in \mathbb{R}^3 \) and \( \textbf{v}_P^{(i)} \in \mathbb{R}^{D_P} \) correspond to atom coordinates and atom features such as element types and amino acid types, respectively. The target output is a binding molecule \( \mathcal{M} = \{(\textbf{x}_M^{(i)}, \textbf{v}_M^{(i)})\}_{i=1}^{N_M} \), where \( N_M \) is the number of atoms in molecule, \( \textbf{x}_M^{(i)} \in \mathbb{R}^3 \) and \( \textbf{v}_M^{(i)} \in \mathbb{R}^{D_M} \). For brevity, we denote \( \textbf{p} = [\textbf{x}_P, \textbf{v}_P] \) (\( \textbf{x}_P \in \mathbb{R}^{N_P \times 3} \), \( \textbf{v}_P \in \mathbb{R}^{N_P \times D_P} \)) and \( \textbf{m} = [\textbf{x}_M, \textbf{v}_M] \) (\( \textbf{x}_M \in \mathbb{R}^{N_M \times 3} \), \( \textbf{v}_M \in \mathbb{R}^{N_M \times D_M} \)) as the concatenation of protein binding site and ligand atoms.

\subsection{Parameter Interpolation Flow}\label{subsec4}
Parameter Interpolation Flow (PIF) is a flow-based generative model that operates in the parameter space of probability distributions. Unlike conventional flow models that transform samples directly, PIF constructs an interpolation path between the parameters of a prior distribution $ \boldsymbol{\theta}_{\text{prior}} $ and a target data-driven distribution $ \boldsymbol{\theta}(\textbf{x}_{\text{data}}) $. Specifically, it learns a time-dependent parameter trajectory $ \boldsymbol{\theta}_t = f(t)\boldsymbol{\theta}(\textbf{x}_{\text{data}}) + (1-f(t))\boldsymbol{\theta}_{\text{prior}} $, where $ f(t) $ is a monotonic function ensuring $ \boldsymbol{\theta}_0 = \boldsymbol{\theta}_{\text{prior}} $ and $ \boldsymbol{\theta}_1 = \boldsymbol{\theta}(\textbf{x}_{\text{data}}) $.  

During training, PIF optimizes the model $ \boldsymbol{\Phi} $ to predict $ \boldsymbol{\theta}(\textbf{x}_{\text{data}}) $ from samples drawn at intermediate $ \boldsymbol{\theta}_t $. The prediction accuracy is evaluated via the KL divergence between the predicted distribution $ p(\textbf{x} \mid \hat{\boldsymbol{\theta}}_{t+\Delta t}) $ and the true interpolated distribution $ p(\textbf{x} \mid \boldsymbol{\theta}_{t+\Delta t}) $. Specifically, $ \boldsymbol{\theta}_{t+\Delta t} $ is the true next-step parameter, calculated as a time-dependent interpolation of the prior and target distributions (Appendix A,B). 
At inference, PIF generates samples through an iterative refinement process: starting from $ \boldsymbol{\theta}_{\text{prior}} $, the model progressively updates parameters until converging to $ \boldsymbol{\theta}(\textbf{x}_{\text{data}}) $(Fig.~\ref{fig1}).

In classical continuous FM framework, the probability path is typically prescribed via linear interpolation in the sample space. While the resulting path corresponds to the optimal transport in the $W_2$ under such linear interpolation when source and target distributions are Gaussian, it fails to maintain such optimality for general distributions. To overcome this, PIF shifts the interpolation mechanism from the sample space to the distribution parameter space, thereby aligning the induced probability path with the intrinsic geometric structure of the distribution space. For continuous variables, PIF characterizes the $W_2$-optimal transport structure between any two one-dimensional distributions of the same type, offering a more flexible and theoretically grounded framework. 

The advantage of PIF is particularly pronounced in discrete variable modeling. Existing discrete FM approaches\citep{gat2024discrete, austin2021structured, zhao2024improving} often rely on $W_2$ distance, which assume an Euclidean structure that categorical variables do not possess. Applying these metrics to the probability simplex is equivalent to imposing a mean-squared-error-type loss, which often fails to capture the true curvature of the manifold—especially near the boundaries—and may lead to distorted interpolation paths and implausible intermediate distributions. In contrast, PIF lifts the problem to a statistical manifold. Under an exponential family prior, interpolating distribution parameters, PIF defines probability paths along exponential geodesics under the Fisher–Rao metric. 

This information-geometric approach offers several advantages: offers a more direct and robust statistical interpretation that better reflects relative changes in probability mass\citep{cheng2024categorical}, ensures insensitivity to the specific choice of parameter representation, and is intrinsically linked to natural gradient dynamics on the statistical manifold, which contributes to improved training stability and efficiency\citep{cheng2024categorical, boll2024generative}. Consequently, PIF provides a unified framework that recovers $W_2$ optimality in the continuous Gaussian case while establishing a metrically consistent and statistically natural foundation for discrete generative modeling.

\subsection{Molecule generation based on PIF}\label{subsec4}
For molecular generation via PIF, we specify the distributions for atomic coordinates $\textbf{x}$ and types $\textbf{v}$ as Gaussian and Dirichlet \citep{stark2024dirichlet}, respectively:
\begin{equation}
p(\textbf{x}) = \mathcal{N}(\textbf{x};\boldsymbol{\mu},\epsilon^2 \textbf{I}) \label{eq8}
\end{equation}
\begin{equation}
p(\textbf{v}) = \text{Dir}(\textbf{v};\boldsymbol{\alpha}) = \frac{1}{B(\boldsymbol{\alpha})} \prod_{i=1}^K \textbf{v}_i^{\boldsymbol{\alpha}_i-1} \label{eq9}
\end{equation}
where $\boldsymbol{\mu} \in \mathbb{R}^3$, $\epsilon$ is a scalar, and $\boldsymbol{\alpha}$ is a $K$-dimensional vector, where $K$ corresponds to the number of atom types, $B(\boldsymbol{\alpha})$ is the multivariate beta function. Accordingly, the parameters of the two distributions are denoted as $\boldsymbol{\theta}_\textbf{x} = (\boldsymbol{\mu},\epsilon)$ and $\boldsymbol{\theta}_\textbf{v} = \boldsymbol{\alpha}$, respectively.

To obtain the parameters corresponding to molecular data, we represent them in the form of Dirac distributions associated with the aforementioned two distribution types. In practice, we extend the definition of Gaussian distributions by setting the standard deviation to zero in the distribution parameters of continuous-variable Dirac distributions:
\begin{equation}
p(\textbf{x}|\textbf{x}_{\text{data}}) = \lim\limits_{\epsilon \to 0^+}\mathcal{N}(\textbf{x};\textbf{x}_{\text{data}}, \epsilon^2 \textbf{I}) \quad \boldsymbol{\theta}_{\textbf{x},\text{data}} = (\textbf{x}_{\text{data}}, 0)
\label{eq10}
\end{equation}
\begin{equation}  
p(\textbf{v}|\textbf{v}_{\text{data}}) = \text{Dir}(\textbf{v};\textbf{v}_{\text{data}}) \quad \boldsymbol{\theta}_{\textbf{v},\text{data}} = \text{Onehot}(\textbf{v}_{\text{data}}) \label{eq11}
\end{equation}
The formulation of the interpolation process is given as follows:
\begin{equation}
\boldsymbol{\theta}_{\textbf{x},t} = f(t) \boldsymbol{\theta}_{\textbf{x},\text{data}} + (1-f(t))\boldsymbol{\theta}_{\textbf{x},\text{prior}} \label{eq12}
\end{equation}
\begin{equation}
\boldsymbol{\theta}_{\textbf{v},t} = f(t) \boldsymbol{\theta}_{\textbf{v},\text{data}} + (1-f(t))\boldsymbol{\theta}_{\textbf{v},\text{prior}} \label{eq13}
\end{equation}
\begin{equation}
\boldsymbol{\theta}_{\textbf{x},\text{prior}} = (\textbf{0},\epsilon_0) \label{eq14}
\end{equation}
\begin{equation}
\boldsymbol{\theta}_{\textbf{v},\text{prior}} = (1/K,1/K,\ldots,1/K) \label{eq15}
\end{equation}
\begin{equation}
f(t) = 1-\gamma^t \label{eq16}
\end{equation}
In the above equation, both $\epsilon_0$ and $\gamma$ are hyperparameters. The formulation of $f(t)$ encourages the model to focus more on learning fine-grained structures in the molecular data, thereby improving the quality of generation.

The loss function is formulated as follows \citep{soch2024statproofbook}:
\begin{equation}
L_{t-\Delta t,\textbf{x}} = \frac{(1-\gamma^t)^2}{2\gamma^t \epsilon_0^2} \mathbb{E}_{p_{\text{data}}} [\|\hat{\boldsymbol{\theta}}_\textbf{x}^{(1)} - \boldsymbol{\theta}_\textbf{x}^{(1)} \|^2] \label{eq17}
\end{equation}
\begin{equation}   
    \begin{split}
        & L_{t-\Delta t,\textbf{v}} = \\ & \mathbb{E}_{p_{\text{data}}} \left[ \ln \frac{\Gamma(\boldsymbol{\theta}_{\mathbf{v},t})}{\Gamma(\hat{\boldsymbol{\theta}}_{\mathbf{v},t})} + (\hat{\boldsymbol{\theta}}_{\mathbf{v},t} - \boldsymbol{\theta}_{\mathbf{v},t})^T (\psi(\hat{\boldsymbol{\theta}}_{\mathbf{v},t}) - \psi(1)) \right]
        \label{eq18}
    \end{split}
\end{equation}

Here, $\Gamma(\textbf{x})$ is the multivariate gamma function, $\psi(\textbf{x})$ is the multivariate digamma function, both $\lambda_\textbf{x}$ and $\lambda_\textbf{v}$ are hyperparameters to adjust the weight of loss.

By conditioning on fixed substructure coordinates and atom types, the model generates the remaining molecular structure: 
\begin{equation}
\boldsymbol{\theta}_{\textbf{x},t,\text{cond}} = \text{Concat}(\boldsymbol{\theta}_{\textbf{x},t},(\textbf{x}_{\text{cond}},0)) \label{eq20}
\end{equation}
\begin{equation}
\boldsymbol{\theta}_{\textbf{v},t,\text{cond}} = \text{Concat}(\boldsymbol{\theta}_{\textbf{v},t},\textbf{v}_{\text{cond}}) \label{eq21}
\end{equation}

MolPIF using Laplace prior is provided in Appendix A.

\subsection{Geometry-enhanced learning strategy}

To improve the capture of atomic-level contextual information, we employ a geometry-enhanced learning strategy based on a dynamic masking mechanism. During training, ligand atoms are stochastically masked (with activation probability $P_m$ and per-atom probability $P_{am}$) and the model is tasked with reconstructing their coordinates and types. This strategy forces the network to act as a structural imputer, learning to respect stringent local geometric constraints such as bond lengths and ring planarity.

\subsection{Implementation details}\label{subsec4}
Atom Featurization: Protein atoms were one-hot encoded by element (H, C, N, O, S, Se) and residue identity, plus a backbone indicator and region tag (arm/scaffold). Ligand atoms were encoded by element (C, N, O, F, P, S, Cl) and aromaticity.\par

Model \& Training: MolPIF's architecture employs the UniTransformer \citep{qu2024molcraft} for equivariance and spatial encoding (See Appendix A for the network structure). Protein-ligand graphs use K-Nearest-Neighbor-based edges. We set $\epsilon_0=1$, $\gamma=0.009$, $\beta_0=1$, the probability of enabling the masking mechanism $P_m=0.3$, the probability of each atom being masked $P_{am}=0.3$, and used 100 sample steps. Training took 24 hours on a single NVIDIA 4090 GPU. For lead optimization, priors were initialized from unfixed region coordinates and atom types.\par

\section{Results}\label{sec2}

\subsection{Experimental setup}

We employed the CrossDocked2020 dataset \citep{francoeur2020three} for training and evaluation. Following the preprocessing protocol AR \citep{luo20213d}, we retained only protein-ligand poses with an RMSD $<$ 1\text{\AA} relative to experimental structures. To ensure generalizability, protein sequences were clustered at 30\% identity using MMseqs2 \citep{steinegger2017mmseqs2}, splitting the data into 99,900 training pairs and 100 test proteins from unseen clusters. For each test protein, 100 molecules were sampled for comprehensive assessment.\par

We evaluate five baselines: AR \citep{luo20213d} generates molecules atom-by-atom via a Markov Chain Monte Carlo \citep{geyer1992practical} method on density grids; Pocket2Mol \citep{peng2022pocket2mol} uses an auto-regressive scheme for 3D positions and bonds; TargetDiff \citep{guan20233d} employs continuous and discrete diffusion for simultaneous generation; DecompDiff \citep{guan2024decompdiff} incorporates scaffold priors and bond diffusion with validity guidance; and MolCRAFT \citep{qu2024molcraft} utilizes the BFNs to bridge the continuous-discrete gap.\par

The performance is assessed via three categories:
(a) Binding Affinity: Calculated via AutoDock Vina \citep{trott2010autodock} using Vina Score (kcal/mol, direct affinity), Vina Min (kcal/mol,post-relaxation affinity), and Vina Dock (kcal/mol,post-redocking affinity).
(b) Chemical Properties: Evaluated by QED \citep{bickerton2012quantifying} (drug-likeness), SA \citep{ertl2009estimation} (synthetic ease), LogP (lipophilicity, target: -0.4 to 5.6 \citep{ghose1999knowledge}), Lipinski score \citep{lipinski1997experimental} (Ro5 compliance), and Diversity (Div, average pairwise Tanimoto distance).
(c) Conformation Stability: Measured by Strain Energy (SE, kcal/mol) \citep{gu2021ligand}, Clash Ratio (CR) via PoseCheck \citep{harris2023benchmarking}, the Jensen-Shannon divergence(JSD) of bond lengths ($\text{JS}_\text{BL}$) and angles ($\text{JS}_\text{BA}$) to quantify structural deviation from references.\par

\subsection{Model Evaluation}\label{subsec2}

\begin{table*}[t]
\caption{
    Comparison of MolPIF and baseline models on the CrossDock test set across 10,000 generated molecules for de novo design. ($\uparrow$) / ($\downarrow$) indicates larger / smaller is better. Top-2 results are highlighted with \textbf{bold} and \underline{underlined}.
}
\centering
\label{tab1}
\resizebox{\linewidth}{!}{%
\begin{tabular}{l|cc|cc|cc|ccc|c|c|c|c|c|cc|c|c}
\toprule
\multirow{2}{*}{Methods}  
& \multicolumn{2}{c|}{Vina Score ($\downarrow$)} 
& \multicolumn{2}{c|}{Vina Min ($\downarrow$)} 
& \multicolumn{2}{c|}{Vina Dock ($\downarrow$)} 
& \multicolumn{3}{c|}{Strain Energy ($\downarrow$)} 
& QED 
& SA 
& LogP 
& Lipinski
& Div 
& \multicolumn{2}{c|}{JS ($\downarrow$)} 
& CR 
& Connected 
\\
& Avg. & Med. 
& Avg. & Med. 
& Avg. & Med.
& 25\% & 50\% & 75\%
& Avg. ($\uparrow$)
& Avg. ($\uparrow$)
& Avg. 
& Avg. ($\uparrow$) 
& ($\uparrow$)  
& BL & BA
& Avg. ($\downarrow$)
& Avg. ($\uparrow$)
\\ \midrule
                            
Reference & -6.36 & -6.46 & -6.71 & -6.49 & -7.45 & -7.26 & 34 & 107 & 196 & 0.48 & 0.73 & 0.89 & 4.27 & - & - & - & 0.17 & - \\ \midrule
AR & -5.75 & -5.64 & -6.18 & -5.88 & -6.75 & -6.62 & 259 & 595 & 2286 & 0.51 & 0.64 & 0.39 & \underline{4.75} & 0.7 & 0.45 & 0.54 & \textbf{0.22} & 0.94 \\
Pocket2Mol & -5.14 & -4.70 & -6.42 & -5.82 & -7.15 & -6.79 & 102 & \underline{189} & \textbf{374} & \underline{0.57} & \textbf{0.76} & 1.51 & \textbf{4.88} & \textbf{0.74} & 0.37 & 0.43 & 0.56 & 0.96 \\
TargetDiff & -5.47 & -6.30 & -6.64 & -6.83 & \underline{-7.80} & \underline{-7.91} & 369 & 1243 & 13871 & 0.48 & 0.58 & 1.36 & 4.51 & 0.72 & \underline{0.26} & 0.48 & 0.53 & 0.90 \\
DecompDiff & -5.19 & -5.27 & -6.03 & -6.00 & -7.03 & -7.16 & 115 & 421 & 1424 & 0.51 & 0.66 & 1.15 & 4.49 & \underline{0.73} & \underline{0.26} & 0.44 & 0.51 & 0.83 \\
MolCRAFT & \underline{-6.55} & \underline{-6.95} & \underline{-7.21} & \underline{-7.14} & -7.67 & -7.82 & \underline{83} & 195 & 510 & 0.50 & 0.67 & 1.16 & 4.46 & \underline{0.73} & \textbf{0.23} & \textbf{0.37} & \underline{0.26} & \underline{0.97} \\
MolPIF & \textbf{-6.64} & \textbf{-7.02} & \textbf{-7.41} & \textbf{-7.28} & \textbf{-8.09} & \textbf{-8.13} & \textbf{65} & \textbf{150} & \underline{375} & \textbf{0.59} & \underline{0.72} & 3.26 & 4.63 & 0.72 & \textbf{0.23} & \underline{0.40} & 0.29 & \textbf{0.98} \\ 

\botrule
\end{tabular}
}
\end{table*}

\subsubsection{Evaluation of common properties for generated molecules}\label{subsec2}

Table \ref{tab1} summarizes the performance of MolPIF compared to baseline models across 10,000 generated molecules. MolPIF consistently demonstrated competitive binding affinity, drug-likeness, and conformational stability.Regarding binding affinity, under size-constrained conditions, MolPIF achieved the lowest mean values in Vina Score (-6.64), Vina Min (-7.41), and Vina Dock (-8.09), outperforming the best AR and TargetDiff baselines by 15.48\% and 21.39\%, respectively. A high Vina Score/Dock ratio (0.82) indicates strong structural consistency between initial generated conformations and docked poses.In terms of chemical properties, MolPIF exhibited superior drug-likeness, yielding the highest QED (0.59) and competitive SA (0.72) and Lipinski compliance. Furthermore, MolPIF showed excellent conformational stability, ranking first in the 25th and 50th percentiles of SE. Low $\text{JS}$ values further confirm its accuracy in capturing local structural distributions. Case study of de novo task is given in Appendix C.

\subsubsection{Analysis of MolPIF on local geometries}\label{subsec2}
We evaluated the local geometry of generated molecules by analyzing ring size frequencies (3–8 membered) and the Jensen-Shannon divergence (JSD) of bond lengths, angles, and torsions (see Appendix G for JSD calculation). As shown in Appendix Table D1, MolPIF effectively avoids unstable small rings (zero 3-membered and only 0.44\% 4-membered rings), while predominantly producing 6-membered rings (76.88\%)—closely matching the reference distribution. In contrast, AR models overproduce small rings, while diffusion-based models struggle to capture the prevalence of 6-membered rings. JSD analysis further confirms that parametric models, particularly MolPIF and MolCRAFT, achieve the lowest divergence from reference distributions, demonstrating superior capability in capturing fine-grained geometric features. Detailed JSD metrics are discussed in Section \ref{sec_prior_dist}.

\subsubsection{Analysis of MolPIF on chemical space distribution}\label{subsec2}
\begin{figure*}[!t]%
\centering
\includegraphics[width=1\textwidth]{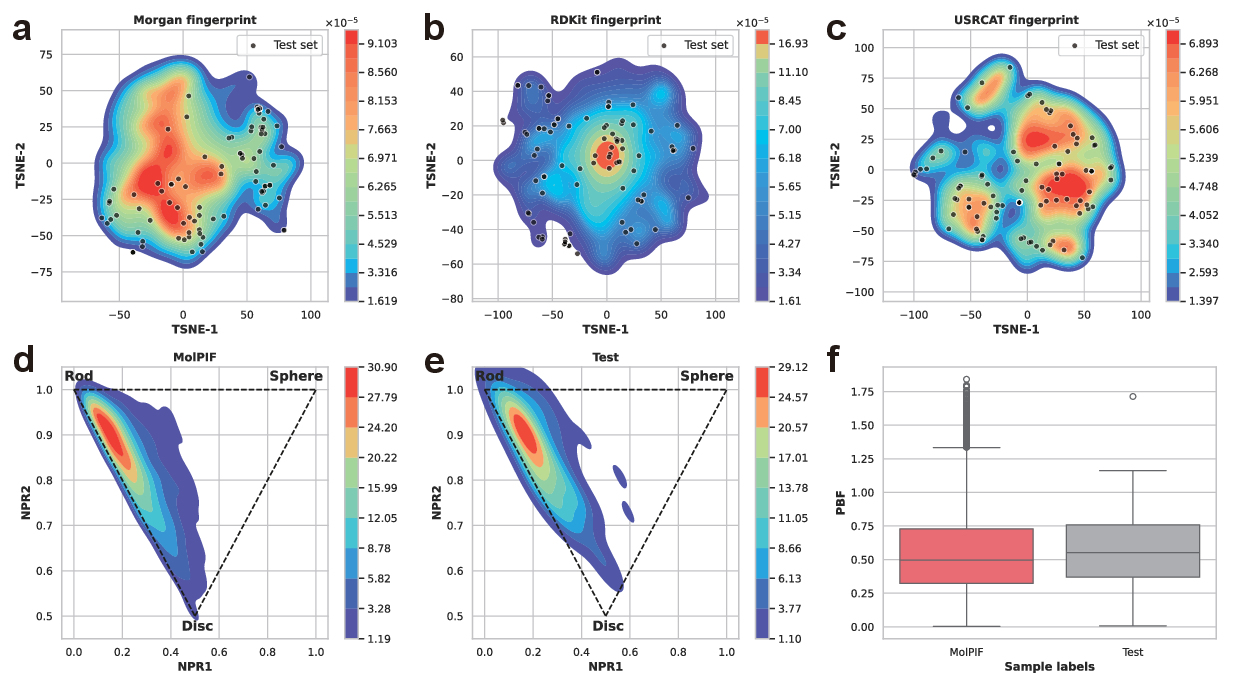}
\caption{\textbf{Chemical space and molecular shape distribution.} \textbf{a-c}, t-SNE visualization based on Morgan (\textbf{a}), RDKit (\textbf{b}), and USRCAT (\textbf{c}) fingerprints. \textbf{d,e}, NPR shape distributions of generated (\textbf{d}) and reference (\textbf{e}) molecules. \textbf{f}, PBF descriptor comparison between generated (n=10,000) and reference (n=100) sets. Box plots indicate median, IQR, 1.5×IQR whiskers, and outliers.
}\label{fig_chemicalspace}
\end{figure*}

To evaluate MolPIF’s performance macroscopically, we analyzed its chemical space distribution using 2D (ECFP, RDKit) and 3D (USRCAT \citep{schreyer2012usrcat}) descriptors (Fig.~\ref{fig_chemicalspace}a–c). While MolPIF-generated molecules comprehensively cover the 2D substructure space of the test set, they exhibit even more precise density alignment in 3D conformational space. This accurate reproduction of high-density regions indicates that MolPIF effectively captures authentic spatial distributions.

Beyond fingerprints, we characterized 3D geometry using Principal Moments of Inertia (PMI \citep{sauer2003molecular}) and Plane of Best Fit (PBF \citep{firth2012plane}). The Normalized Principal Moment of Inertia (NPR) ternary plot (Fig.~\ref{fig_chemicalspace}d,e) shows that MolPIF not only matches the test set’s concentration near rod-shaped morphologies but also explores underrepresented disc- and sphere-shaped regions. Furthermore, the PBF distribution (Fig.~\ref{fig_chemicalspace}f) shows strong agreement between MolPIF-generated molecules and the test set, confirming the model's ability to reproduce spatial planarity. Overall, MolPIF balances fidelity to structural trends with the exploration of novel molecular shapes.

\subsubsection{Analysis of the prior distribution selection}\label{sec_prior_dist}
We explored the flexibility of prior distributions by replacing the Gaussian distribution in MolPIF with a Laplace distribution (MolPIF(La)). As shown in Table \ref{tab2}, while the Gaussian prior generally yields superior binding affinities (e.g., higher Vina scores), the Laplace prior excels in chemical properties (QED, LogP), molecular diversity (0.75), and substructure accuracy. Notably, MolPIF(La) achieved the lowest JSD in bond length distributions and captured distinct modes that other models missed (Appendix Fig. D1, Tables D2-D4).The mask module exhibited divergent effects: in the Gaussian model, it prioritized conventional properties over substructure precision; conversely, in the Laplace model, it significantly enhanced both binding metrics and substructure generation. Despite these trade-offs, MolPIF(La) still outperformed most baselines in Table \ref{tab1}

These phenomena stem from the interaction between the prior's inductive bias and the mask module. While the Gaussian prior favors global smoothness, it conflicts with the local discontinuities introduced by masking, leading to over-smoothed reconstructions. In contrast, the Laplace prior, with its heavy tails, aligns with the local sparsity of molecular substructures. This synergy enhances the model’s ability to reconstruct masked regions with high fidelity. Consequently, MolPIF with a Laplace prior achieves superior structural realism compared to the Gaussian baseline. Furthermore, MolPIF maintains strong performance with non-conventional priors without the complex derivations required by diffusion models, offering significant flexibility for SBDD tasks.

\begin{table*}[t]
\caption{Comparison of MolPIF variants on the CrossDock test set across 10,000 generated molecules for de novo design. ($\uparrow$) / ($\downarrow$) indicates larger / smaller is better. Top-2 results are highlighted with \textbf{bold} and \underline{underlined}.
}
\centering
\label{tab2}
\resizebox{\linewidth}{!}{%
\begin{tabular}{l|cc|cc|cc|ccc|c|c|c|c|c|cc|c}
\toprule
\multirow{2}{*}{Methods}  
& \multicolumn{2}{c|}{Vina Score ($\downarrow$)} 
& \multicolumn{2}{c|}{Vina Min ($\downarrow$)} 
& \multicolumn{2}{c|}{Vina Dock ($\downarrow$)} 
& \multicolumn{3}{c|}{Strain Energy ($\downarrow$)} 
& QED 
& SA 
& LogP 
& Lipinski
& Div 
& \multicolumn{2}{c|}{JS ($\downarrow$)} 
& CR \\
& Avg. & Med. 
& Avg. & Med. 
& Avg. & Med.
& 25\% & 50\% & 75\%
& Avg. ($\uparrow$)
& Avg. ($\uparrow$)
& Avg. 
& Avg. ($\uparrow$) 
& ($\uparrow$)  
& BL & BA
& Avg. ($\downarrow$)
\\ \midrule
                            
MolPIF & \underline{-6.64} & \textbf{-7.02} & \textbf{-7.41} & \textbf{-7.28} & \textbf{-8.09} & \textbf{-8.13} & \textbf{65} & \textbf{150} & \textbf{375} & \textbf{0.59} & \textbf{0.72} & 3.26 & \underline{4.63} & 0.72 & 0.23 & \underline{0.40} & \underline{0.29} \\ 
MolPIF(w/o mask) & \textbf{-6.78} & \underline{-6.99} & \underline{-7.28} & \underline{-7.18} & \underline{-7.90} & \underline{-7.95} & \underline{73} & \underline{173} & 467 & 0.55 & 0.70 & 2.28 & 4.49 & 0.72 & \underline{0.20} & \textbf{0.39} & \textbf{0.24} \\ 
MolPIF(La) & -5.17 & -6.02 & -6.43 & -6.70 & -7.64 & -7.76 & 77 & 203 & 691 & 0.55 & 0.70 & 2.45 & 4.51 & \underline{0.73} & \textbf{0.16} & 0.40 & 0.39 \\ 
MolPIF(La w/o mask) & -4.75 & -6.10 & -6.28 & -6.68 & -7.84 & -7.90 & 81 & 183 & \underline{417} & \underline{0.56} & \textbf{0.72} & 3.42 & \textbf{4.65} & \textbf{0.75} & 0.26 & 0.47 & 0.36 \\ 

\botrule
\end{tabular}
}
\end{table*}

\subsubsection{Generalization Ability of MolPIF}
To rigorously evaluate model generalization, we adopt PoseBusters \citep{buttenschoen2024posebusters} as a challenging out-of-distribution (OOD) test set. Unlike the widely used CrossDock dataset, which contains noisy software-generated poses, PoseBusters consists exclusively of high-quality, wet-lab crystal structures released since 2021. To prevent information leakage, we follow \citep{qiu2025piloting} and employ MMseqs2 to discard any test proteins sharing $>$30\% sequence identity with the training set, resulting in a rigorously curated subset of 180 test proteins. For each test protein, we generate 100 molecules using MolPIF and baseline models.

\begin{figure}[!t]%
\centering
\includegraphics[width=\linewidth]{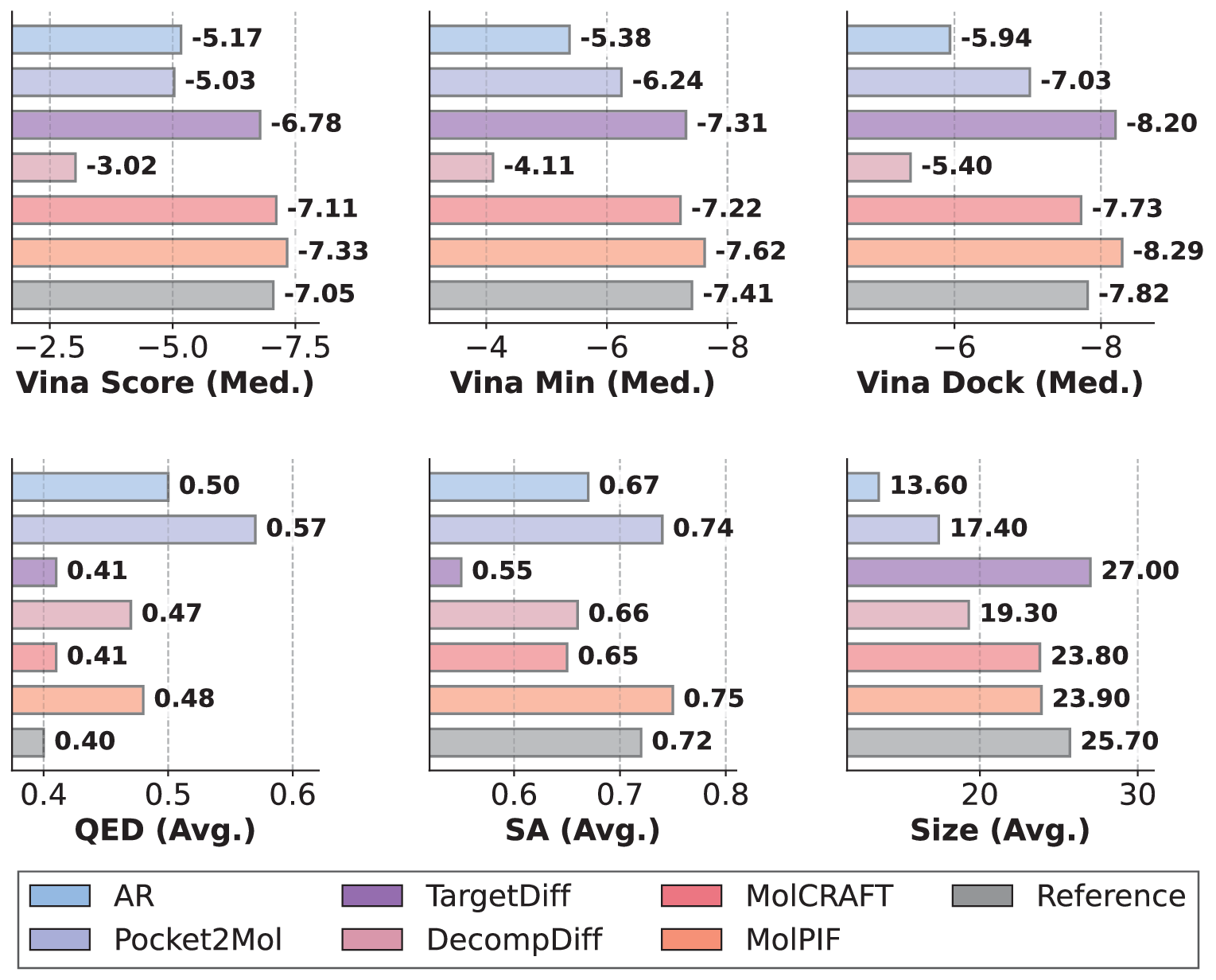}
\caption{
Comparison of MolPIF and baseline models on the PoseBusters OOD test set across 18,000 generated molecules for de novo design.
}
\label{pb_res}
\end{figure}

As shown in Fig. \ref{pb_res}, MolPIF maintains robust performance within this OOD setting. Regarding binding affinity, drug-likeness, and synthetic accessibility, the molecules generated by MolPIF even surpass the reference molecules. This level of performance was not achieved by any of the baselines in the OOD scenario, making MolPIF the sole model to do so in this experiment. Furthermore, the molecular size data confirms that the superior binding affinity is driven by more rational structural design rather than a mere increase in atom count. These results demonstrate the strong generalization ability of MolPIF, which effectively captures underlying structure-activity relationships and generates high-quality molecules even in challenging OOD scenarios.

\subsubsection{Ablation study}

Table \ref{tab:ablation_study} illustrates the effectiveness of each component within the MolPIF framework. We evaluate the choice of the time-dependent function $f(t)$ within our formulation $\theta_t = f(t)\theta_{data} + (1-f(t))\theta_{prior}$. Comparing linear ($f(t)=t$), quadratic ($f(t)=t^2$), and exponential ($f(t)=1-\gamma^t$) schedules, empirical results show that $f(t)=1-\gamma^t$ yields the best performance. This suggests that a convex $f(t)$ is advantageous, as it reduces the proportion of the high-noise stage and helps the model capture more accurate molecular structures. Furthermore, a comparison of different decay rates ($\gamma \in \{0.005, 0.009, 0.02\}$) confirms that $\gamma=0.009$ is the optimal setting for our model.

To isolate the improvements introduced by PIF, we compare it against standard flow matching, which typically employs a Gaussian distribution with linear interpolation for atomic coordinates and a categorical distribution with linear interpolation for atom types. Using the identical UniTransformer architecture and training data, PIF clearly outperforms the standard flow matching baseline. Additionally, comparative experiments on toy datasets (see Appendix B) consistently demonstrate PIF's superiority over common generative frameworks under strictly controlled settings, further validating our architectural design.

\begin{table}[!h]
\caption{Ablation studies of MolPIF on the CrossDock test set across 10,000 generated molecules for de novo design. ($\uparrow$) / ($\downarrow$) indicates larger / smaller is better. Top-2 results are highlighted with \textbf{bold} and \underline{underlined}. Note: V.S. = Vina Score, V.M. = Vina Min, V.D. = Vina Dock, Q. = QED, S. = SA, S.E. = Strain Energy, F.M. = Flow Matching.
}
\centering
\label{tab:ablation_study}
\resizebox{\linewidth}{!}{%
\begin{tabular}{cl|cccccc} 
\toprule
\multicolumn{2}{l|}{Methods} & $f(t)=t$   & $f(t)=t^2$ & $\gamma=0.02$ & $\gamma=0.005$ & F.M. & Best \\
\midrule
\multirow{2}{*}{\rotatebox[origin=c]{90}{V.S.}} & Avg. ($\downarrow$) & \textbf{-6.72} & -6.54 & -6.34 & -6.1 & -5.66 & \underline{-6.64} \\
& Med. ($\downarrow$) & \underline{-6.84} & -6.51 & -6.72 & -6.53 & -6.3 & \textbf{-7.02} \\ 
\midrule
\multirow{2}{*}{\rotatebox[origin=c]{90}{V.M.}} & Avg. ($\downarrow$) & \underline{-7.18} & -6.93 & -6.98 & -6.89 & -6.67 & \textbf{-7.41} \\
& Med. ($\downarrow$) & \underline{-7.06} & -6.7  & -6.96 & -6.92 & -6.73 & \textbf{-7.28} \\ 
\midrule
\multirow{2}{*}{\rotatebox[origin=c]{90}{V.D.}} & Avg. ($\downarrow$) & \underline{-7.89}     & -7.70     & -7.78     & -7.68     & -7.67     & \textbf{-8.09}     \\
& Med. ($\downarrow$) & \underline{-7.97}     & -7.70     & -7.77     & -7.87     & -7.81     & \textbf{-8.13}     \\ 
\midrule
\multirow{1}{*}{\rotatebox[origin=c]{90}{Q.}} & Avg. ($\uparrow$) & 0.55 & 0.48 & \underline{0.57} & \underline{0.57} & 0.54 & \textbf{0.59} \\ 
\midrule
\multirow{1}{*}{\rotatebox[origin=c]{90}{S.}} & Avg. ($\uparrow$) & 0.69 & 0.64 & \underline{0.74} & \textbf{0.78} & 0.71 & 0.72 \\ 
\midrule
\multirow{3}{*}{\rotatebox[origin=c]{90}{S.E.}} & 25\% ($\downarrow$) & 99.62 & 229.31 & 81 & \textbf{64} & 91 & \underline{65} \\
& 50\% ($\downarrow$) & 369.94 & 758.55 & 232 & \textbf{148} & 216 & \underline{150} \\
& 75\% ($\downarrow$) & 1247.60 & 2469.64 & 718 & \textbf{363} & 514 & \underline{375} \\
\botrule
\end{tabular}
}
\end{table}

\subsubsection{The performance of MolPIF in lead optimization}

\begin{figure*}[!t]%
\centering
\includegraphics[width=0.9\textwidth]{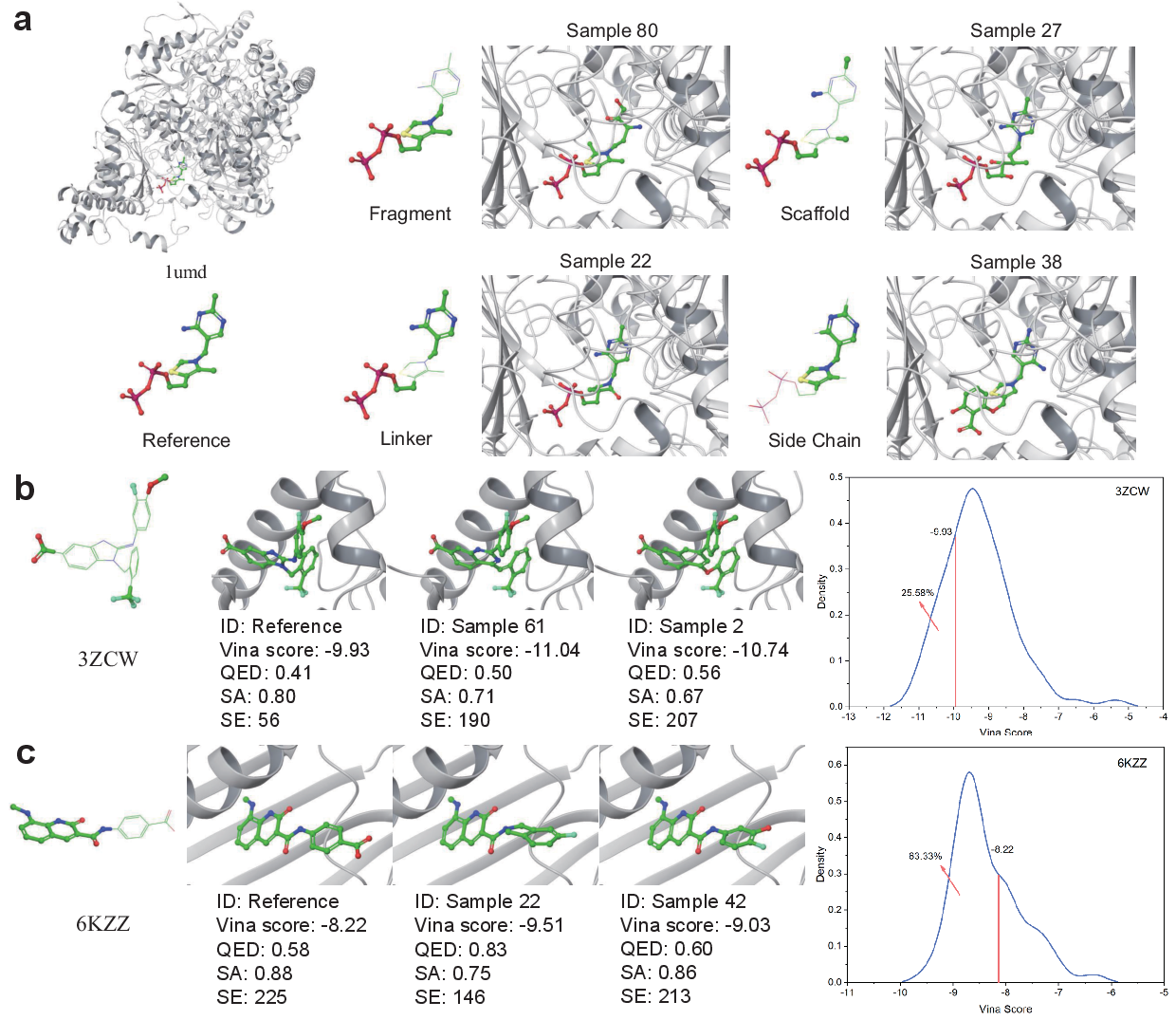}
\caption{
\textbf{Case study of generated molecules in lead optimization scenarios.} 
\textbf{a}, 3D structures of selected MolPIF-generated molecules for target 1umd across four optimization scenarios.
\textbf{b},\textbf{c}, 3D structures of MolPIF-generated molecules for targets 3ZCW (scaffold hopping) and 6KZZ (fragment growth) alongside reference molecules, accompanied by their respective Vina score distributions.
}
\label{fig5}
\end{figure*}

MolPIF enables atom-level lead optimization by allowing users to fix specific atoms while generating complementary substituent regions. 
We validated this capability using targets 1umd, 3ZCW\citep{alexandar2022coarse}, and 6KZZ\citep{ushiyama2020lead}. 
For 1umd(Fig.~\ref{fig5}a), following the CBGBench\citep{lin2024cbgbench} protocol (see Appendix F), MolPIF demonstrated versatility in both extending scaffold groups (fragment/side-chain modification) and integrating discrete fragments (linker / scaffold operations). 
This atom-level control facilitates the rapid modification of physicochemical properties. 
For 3ZCW, we adopted the same side-chain fixation strategy as Delete \citep{chen2025deep}. Alternatively, for 6KZZ we relaxed the atomic constraints compared to DeepFrag \citep{green2021deepfrag} to better demonstrate lead optimization potential. 
100 optimized molecules have been generated for each target.
In scaffold hopping tasks for 3ZCW(Fig.~\ref{fig5}b), 25.58\% of MolPIF-optimized ligands surpassed the reference Vina score ($-9.93$). In fragment growth tasks for 6KZZ(Fig.~\ref{fig5}c), 63.33\% of generated molecules outperformed the reference score ($-8.22$). Across all subtasks, MolPIF consistently produced a higher proportion of molecules with superior binding affinity and 3D spatial compatibility compared to  Delete and DeepFrag (Appendix Tables D5-D6). 
These results underscore MolPIF's broad applicability and potential to enhance lead optimization.

\section{Conclusion}\label{sec3}
We introduce Parameter Interpolation Flow (PIF), a novel generative framework that operates in the parameter space of probability distributions, enabling smooth transformations between prior and target distributions for both continuous and discrete data. PIF overcomes the limitations of traditional flow models, diffusion models, and Bayesian flow networks by interpolating between distributions in the parameter space rather than sample space.

Theoretically, we demonstrate that PIF recovers $W_2$-optimal transport for continuous variables and establishes Fisher–Rao geodesics for discrete data under exponential family priors, preserving the intrinsic manifold structure.

We apply PIF in MolPIF for SBDD. Our empirical evaluations on the CrossDocked2020 dataset confirm that MolPIF achieves superior performance in generating molecules with high chemical validity, accurate geometries, and strong binding affinities. Chemical space analysis confirms the model’s ability to replicate reference distributions and expand structural diversity beyond known chemical territories.
Our analysis of prior distributions (Gaussian vs. Laplace) further illuminated the impact of prior selection on generation quality and structural refinement.
MolPIF also shows promising results in lead optimization, generating novel candidates with enhanced docking scores and preserved critical binding characteristics.

Despite its efficacy, MolPIF has several limitations. First, it treats the protein binding pocket as a rigid structure, omitting the induced-fit dynamics crucial in physiological environments. Second, the framework currently lacks explicit attribute-guided mechanisms for fine-grained lead optimization. Lastly, although MolPIF is computationally efficient, further reducing the sampling steps through advanced solvers or distillation is necessary for its application in ultra-large-scale virtual screening.

In summary, PIF offers a new paradigm for generative modeling in drug discovery. By bridging the gap between heterogeneous modalities within a unified theoretical framework, MolPIF enables a more consistent and accurate exploration of the vast chemical space. Future work will explore extending PIF to more complex distribution priors and broader applications in molecular and materials science, as well as the attribute-guided mechanism under the MolPIF framework.

\section{Data availability}
Our model weights, configuration files, and generated molecules are publicly available at \url{https://drive.google.com/drive/folders/1VBGnHyThNHpdaLgppOeKCKomwfL6oXde}.
The code of MolPIF is freely available at \url{https://github.com/BLEACH366/MolPIF}, and is archived at Zenodo: \url{https://doi.org/10.5281/zenodo.16925025}.

\section{Acknowledgments}
This work was supported by Shanghai Rising-Star Program (23QD1400600) and National Key Research and Development Program of China (2022YFC3400504). 

\section{Competing Interests Statement}
The authors declare that they have no competing interests.


\bibliographystyle{oup-abbrvnat}
\bibliography{molpif}


\end{document}